\documentclass[runningheads]{llncs}

\usepackage{graphicx}

\usepackage{amssymb}
\usepackage{supertabular}
\usepackage{amsmath}
\usepackage{algpseudocode}
\usepackage{algorithm}
\usepackage{tabularx}
\usepackage{multicol}
\usepackage{lineno}
\usepackage{booktabs}
\usepackage{lscape}
\usepackage{color}
\usepackage{subfigure}
\begin{document}
\title{Fixed set search applied to the traveling salesman problem}
%
%\titlerunning{Fix set search applied to the traveling salesman problem}
% If the paper title is too long for the running head, you can set
% an abbreviated paper title here
%

\author{Raka Jovanovic\inst{1} \and
Milan Tuba\inst{2} \and
Stefan Voss\inst{3,4}}
\authorrunning{R. Jovanovic et al.}
% First names are abbreviated in the running head.
% If there are more than two authors, 'et al.' is used.
%
\institute{Qatar Environment and Energy Research Institute (QEERI), Hamad bin Khalifa University,
              PO Box 5825, Doha, Qatar
              \and
              Department of Technical Sciences, State University of Novi Pazar, Vuka Karadzica bb, 36300 Novi Pazar, Serbia
               \and
Institute of Information Systems, University of Hamburg, Von-Melle-Park 5, 20146 Hamburg, Germany
               \and
               Escuela de Ingenieria Industrial, Pontificia Universidad Cat\'olica de Valpara\'iso, Chile\\               
\email{rjovanovic@hbku.edu.qa}
}
\maketitle              % typeset the header of the contribution
\begin{abstract}
In this paper we present a new population based metaheuristic called the fixed set search (FSS). The proposed approach represents a method of adding a learning mechanism to the greedy randomized adaptive search procedure (GRASP). The basic concept of FSS is to avoid focusing on specific high quality solutions but on parts or elements that such solutions have. This is done through  fixing a set of elements that exist in such solutions and dedicating computational effort to finding near optimal solutions for the underlying subproblem. The simplicity of implementing the proposed method is illustrated on the traveling salesman problem. Our computational experiments show that the FSS manages to find  significantly better solutions than the GRASP it is based on and also the dynamic convexized method. 

\end{abstract}
\section{Introduction}
In the last several decades there has been an extensive research effort on developing different metaheuristics for finding near optimal solutions for hard optimization problems. Most metaheuristic approaches focus on how to balance the global search (exploration) and  local search (exploitation) in examining the solution space.  There have been several directions in this research. Early methods include simulated annealing \cite{Sim1} and tabu search \cite{Tabu1,Tabu2} where the search is focused near the best found solution and on mechanisms of escaping local optima. In later stages population based methods have proven to be very powerful. The general approach in such methods is generating a large number of solutions and including different types of learning mechanism. In case of genetic algorithms \cite{GA} and differential evolution \cite{DIF} the main idea is in combining different high quality solutions with the addition of a certain level of randomization. Particle swarm optimization \cite{PSO1,PSO2}   explores the solution space through generating new solutions based on the positions of the globally and locally best found solution. This basic idea has been incorporated in a wide range of similar methods like cuckoo search \cite{CS}, artificial bee colony algorithm \cite{ABC}, bat algorithm \cite{yang2013bat}, krill herd algorithm  \cite{GANDOMI20124831} and many others. (The reader should note that we are fully aware about the controversial discussion with respect to some of these or similar approaches; see \cite{SoerensenITOR}.) The ant colony optimization \cite{dorigo2005ant,Jovanovic20115360} uses a population based method to add a learning mechanism to greedy algorithms.

One of the most common methods for improving population based metaheuristics is  by combining them   with  local searches. The variable neighborhood search \cite{hansen2001variable} metaheuristic focuses on the efficient use of local searches. The performance of the original metaheuristics is often improved by different types of enhancements or by creating hybridized methods that combine one or more of such metaheuristic methods \cite{blum2011hybrid,TSAI200467,MARINAKIS20114684}. The main problems with such methods is the increased complexity of implementation. This problem is most evident if we observe publication in fields other than operational research and applied mathematics. In the vast majority of them only the original, simple to implement, method is used to solve the problem of interest. 

The Greedy randomized adaptive search procedure (GRASP) \cite{feo1995greedy,hs87} is one example of how simplicity is important. Although it generally has a worse performance than combining one of the more complex methods with a local search it is extensively used. The advantage of more complex metaheuristics often occurs only for very large problem instances, some examples in case of ACO can be seen in \cite{jovanovic2016ant,jovanovic2018partitioning}. Because of this, it is reasonable to attempt to increase  the size of problems that GRASP can solve, but in a way that there is a  minimal increase in complexity of the original method.

In this paper we focus on developing this type of method through adding a simple learning mechanism to GRASP. Some examples of such methods are GRASP with path relinking \cite{Festa2013} and the dynamic convexized method    		\cite{7984713}. Both of these methods produce a significant level of improvement. Let us note that both of them  use the standard concept of intensifying  the search of the solution space based on the location of globally and locally best found solutions. The basic concept of the proposed  fixed set search method (FSS) is to avoid focusing on specific high quality solutions but on parts or elements that such solutions have. This idea of exploiting elements that belong to high quality solutions is used in ACO, were the randomized greedy algorithm is directed to chose such elements. 
The ideas for developing  this method may be based on earlier notions of chunking \cite{vg98,woo98}, vocabulary building and consistent chains \cite{sv99} as they have been used, e.g., in relation to tabu search. In those notions one relates given solutions of an optimization problem as
composed of parts (or chunks). Considering the traveling salesman problem (TSP), for instance, a part may be a set of nodes to be visited consecutively. Moreover, some parts may be closely related to some other parts so that a corresponding connection can be made between two parts. Similar ideas are even found in the POPMUSIC paradigm \cite{TaillardVoss99}.
 The general idea of the proposed approach is  to fix a set of elements that exist in high quality solutions and dedicate computational effort on ``filling in the gaps".

 This concept is illustrated on the symmetric TSP through adding a learning mechanism to GRASP.  We should note that exact codes for the TSP are available \cite{CONCORDE}. Nevertheless, due to its widespread investigation it seems appropriate to use it for illustration purposes. As it will be seen in the following, this type of approach can easily be added to existing GRASP algorithms and produces a high level of improvement in the quality of found solutions and computational cost.

The paper is organized as follows. In the first section we give a brief description of the GRASP method for the TSP. In the following section we present the FSS and show how it is applied to the TSP. In Section 4, we discuss the  performed computational experiments.

\section{GRASP}

In this section we give a short outline of the GRASP algorithm used for solving the TSP. The pseudocode for the general GRASP algorithm is given in Alg. \ref{Alg:GRASP}.

\begin{algorithm}
\begin{algorithmic}
%\STATE{BFSGrowBiConnected(Graph, r)}
\While{ Not Stop Criteria Satisfied}
\State{Generate Solutions $S$ using randomized greedy algorithm}
\State{Apply local search to $S$}
\State{Check if $S$ is the new best}
\EndWhile
\end{algorithmic}
\caption{\label{Alg:GRASP} Pseudocode for GRASP}
\end{algorithm}

In the case of the TSP it is common to use a randomization of the nearest neighbor greedy algorithm with a  restricted candidate list (RCL) based on the cardinality of nodes \cite{RCLDLB}. In case of the local search, the most commonly used ones are the 2-OPT \cite{2OPT} and 3-OPT \cite{3OPT} searches. In practice instead of the original versions of the two local searches it is common to use a RCL of edges that will be used for evaluating the proposed improvement. 

\section{Fixed set search}

In this section we present the proposed fixed  set search metaheuristic and show how it can be used in combination with GRASP. Before giving the details of the method we give the basic concepts on which it is constructed.

One of the main disadvantages of GRASP is the fact that it does not incorporate any learning mechanism. On the other hand, such an improvement should be designed in a way that it is simple to implement. In this paper we propose one such method called the fixed set search (FSS). In the following  we will assume that a solution $S$ of the problem of interest can be represented in the form of a set. In case of the TSP, the solution $S$ can be viewed as a set of edges $\{ e_1, e_2, .., e_2\}$. The development of FSS is based on two simple premises:

\begin{itemize}
\item{A combinatorial optimization problem is generally substantially easier to solve if we fix some parts of the solution, and in this way lower the size of the solution space that is being explored.}
\item{There are some parts of high quality solutions that are ``easy to recognize". We say this in the sense that they appear in many good solutions. In general there is no need to dedicate a significant amount of computational effort to analyze them. }
\end{itemize}

The general idea of FSS is to fix such ``easy to recognize" parts of good solutions and dedicate computational effort in finding the optimal solution for the corresponding subset of the solution space.  Informally, we take the common sections of good solutions, which we will call the {\it fixed set}, and try to ``fill in the gaps". 
In the following sections we will illustrate how this simple idea can be incorporated in the GRASP metaheuristic for the TSP.  The proposed algorithm has three basic steps. The first one is finding a fixed set. The second is adapting the randomized greedy algorithm to be able to use a preselected set of elements. Finally, specify and apply the method which gains experience from previously generated solutions.

\subsection{Fixed set}

As previously stated, to be able to implement the proposed method is is necessary that we can represent a solution of the problem in the form of a set $S$. In case of the TSP, a solution $S$ corresponds to the set of edges that represent a Hamiltonian cycle.  Let us use the notation $\mathcal{P}$ for the set of all the generated solutions (population). In relation, let us define $\mathcal{P}_n$ as the set of $n$ best generated solutions based on the objective function, the path length. Further, let us use the notation $F$ for a {\it fixed set} that will be used in the search. In the following we define a method for finding a {\it fixed set} $F$ for a population of solutions $\mathcal{P}$. The proposed method should satisfy the following requirements:
\begin{itemize}
\item{(R1) A generated fixed set $F$ should consist of elements of high quality  solutions.}
\item{(R2) The method should be able to generate  many different random fixed sets  that can be used to generate new high quality solutions.}
\item{(R3) A generated fixed set $F$ can be used to generate a feasible solution. More precisely, there exists a feasible solution $S$ such that $F \subset S$} 
\item{(R4) Ability to control  the size of the generated fixed set $|F|$.}
\end{itemize}

The first two requirements can be achieved if we  only use some randomly selected high quality solutions for generating the fixed sets. This can be achieved by simply selecting $k$ random solutions from the set  $\mathcal{P}_n$.  Let us define $\mathcal{S}_{kn}$ as the set of selected solutions. The initial idea is to use the intersection of all the solutions in $\mathcal{S}_{kn}$ for the fixed set $F$. The problem is that we have no control over the size of the intersection. A simple idea to control the size of $F$ is instead of using the intersection of $\mathcal{S}_{kn}$, is to select the elements (edges) that are a part of the highest number of solutions. The problem with this approach is that such a selection can potentially contain edges that could not be used to generate a feasible solution.

Both of these issues can be avoided if a base solution $B \in \mathcal{P}_m$ is used in generating a fixed set $F$.  More precisely, we can select the elements of $B$ that occur most frequently in $\mathcal{S}_{kn}$. Let us define this procedure more formally. We will assume that we are finding a fixed set $|F| = Size$ for a set of solutions $\mathcal{S}_{kn} = \{S_1, .., S_k\}$ and base solution $B = \{ e_1, ... e_m\}$. Let use define the function $C(e_x, S)$ which is equal to $1$ if $e_x \in S$ and $0$ otherwise. Using $C(e_x, S)$ we can define the  function that counts the number of times an edge $e_x$ occurs in the $\mathcal{S}_{kn}$ as follows.

\begin{equation}
O(e_x, \mathcal{S}_{kn}) = \sum_{S \in \mathcal{S}_{kn}}C(e_x,S)
\end{equation}

\begin{figure}[thb]
\centering
\includegraphics[width=0.85\textwidth]{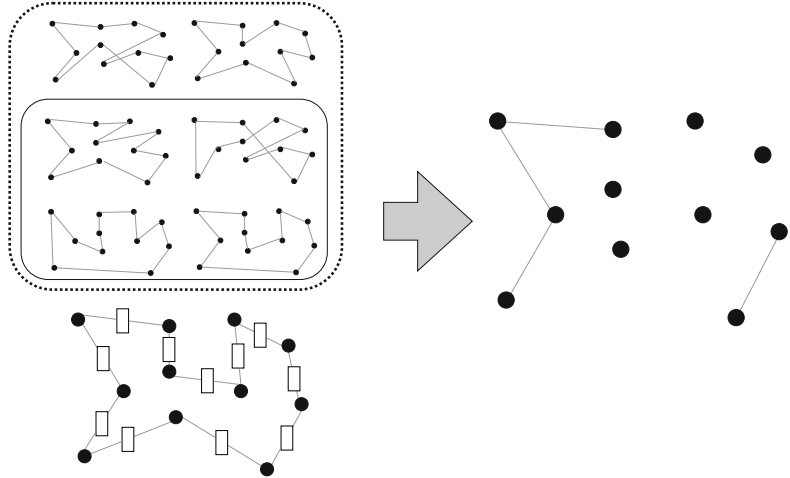}

\caption{\label{fig:FixSet} Illustration of generating a fixed set. The input is $\mathcal{S}_{kn}$ (top left), a set of four randomly selected solutions out of the six best ones, and a base solution $B$ (left bottom). The value on an edge of $B$ represent the number of occurrences of that edge in elements of $\mathcal{S}_{kn}$. The edges on the right hand side present the corresponding fixed set of size four.}
\end{figure}

Now,  we can define $F \subset B$ as the set of edges $e_x$ that have the largest value of $O(e_x, \mathcal{S}_{kn})$. In relation let us define function $F = Fix(B,\mathcal{S}_{kn}, Size)$ that corresponds to the fixed set generated for a base solution $B$, a set of solutions $\mathcal{S}_{kn}$ having $Size$ elements.  An illustration of the method for generating a fixed set for the TSP can be seen in Fig. \ref{fig:FixSet}.

\subsection{Randomized greedy algorithm with preselected elements}

To be able to use the fixed set within a GRASP setting we need to adapt the greedy randomized algorithm. Let us first note that in case of the TSP, the fixed set $F$ will consist of several paths (sequence of edges which connect a sequence of vertices) of graph $G$.  This effects the greedy algorithm in two ways.  First, the inside nodes of the path should be removed from the candidate list.  Secondly, if a node that is a start or end node of a path in the fixed set, is added to the current partial solutions the whole path must be added in the proper direction. Pseudocode for the adapted greedy algorithm can be seen in the Alg. \ref{Alg:RGF}. 

\begin{algorithm}
\begin{algorithmic}
\State{Set $Paths$ to all paths in $F$}
\State{$Candidates = V$}
\State{$Candidates = Candidates \setminus F$}
\ForAll{$p \in Paths$}
\State $Candidates = Candidates \cup p[First] \cup p[Last]$
\EndFor
\State{Select Random start city from $Candidates$}
\While{Not Completed Tour}
\State{Select next city $c$ using $RCL$}
\ForAll{$p \in Paths$}
\If {$(c = p [First]) \wedge (c = p [Last])$} 
\State{Add path $p$ to current solution in correct direction}
\State{{\textbf{break}}}
\EndIf
\EndFor
\EndWhile
\end{algorithmic}
\caption{\label{Alg:RGF} Pseudocode for Greedy algorithm with preselected elements}
\end{algorithm}

In relation, let us define the function $S = RGF(F)$, for a fixed set $F$, as the solution acquired using this algorithm. 

\subsection{Learning mechanism}

In this section we present the FSS which is used as a learning mechanism for GRASP.  Before presenting details of the proposed methods, let us first make a few observations. In the general case the early iterations of GRASP frequently manage to improve the quality of the best found solution. At later stages such improvements become significantly less frequent and the method becomes dependent on ``lucky" hits. The idea is to use a fixed set $F$, for some promising region of the solution space, generate a solution $S = RGF(F)$ and apply a local search to $S$.  In this way we increase the probability that a higher quality solution will be found. An important aspect is how to select the size of the fixed set. In case it is small, it efficiently performs a global search but after a certain number of executions, as in the case of GRASP, it will to a large extend be dependent on ``lucky" hits. On the other hand if $F$ is large, it will only explore the parts of the solution space that are close to already generated solutions. As a consequence, there is a high risk of being trapped in locally optimal solutions.

This indicates that the size of the fixed set should be adapted during the execution of the algorithms.  For simplicity, we can a priori define an array $Sizes$  of fixed set sizes that will be tested, using the following formula:

\begin{equation}
\label{FixedSizes}
 Sizes[i] = |V| - \frac{V}{2^i}
\end{equation} 

In \eqref{FixedSizes}, $V$ represents the set of nodes of the graph on which the TSP is being solved. The maximal value of an element in the array $Sizes$ is chosen based on the problem being solved. Using this array let us give an outline of the FSS. We will first generate an initial population of solutions $\mathcal{P}$ by executing GRASP for $N$ iterations. This initial population will be used to find the fixed sets. We start from a small fixed set and generate solutions until stagnation, in the sense of not finding new best solutions for a large number of iterations, has occurred. When stagnation occurs, we increase the size of the fixed set (selecting the next element of $Sizes$) and repeat the procedure. In this way  a more focused exploration of the search space is executed. This procedure is repeated until the largest element in $Sizes$ is tested. 

At this stage it is expected that the set $\mathcal{P}_m$ of $m$ best solutions has significantly changed and contains higher quality solutions than in the initial population. Because of this,  there is a potential that even for smaller sized fixed sets, since they are now generated using better solutions,  there is a higher probability of finding new quality solutions. So, we can repeat the same procedure from the smallest fixed set size. Let us note that after a large number of solutions is generated the new solutions acquired using small fixed sets are rarely new best ones. The importance of their revisit  is in generating new types of high quality solutions. In case not even high quality solutions, not being one of the  $m$ best ones, can be found, small fixed sets can be excluded  from the search. The pseudocode for FSS can be seen in Alg. \ref{Alg:FSS}.

\begin{algorithm}
\begin{algorithmic}

\State{Initialize $Sizes$}
\State $Size = Sizes.Next$
\State Generate initial population  $\mathcal{P}$ using $GRASP(N)$
  \State
 \While{(Not termination condition)}
 \State
 \State{Set $\mathcal{S}_{kn}$ to random $k$ elements of $\mathcal{P}_n$}		
 \State{Set $B$  to a random solution in $\mathcal{P}_m$}		
 \State{$F = Fix(B, \mathcal{S}_{kn}, Size)$}
 \State
  \State{$S = RGF(F)$}
  \State{Apply local search to $S$}
  \State{$\mathcal{P} = \mathcal{P} \cup \{S\}$}
   \State
 \If{Stagnant Best Solution}
  \If{(Stagnant Candidates) $\wedge$ ($Size = Min(Sizes)$)}
		 \State	{Remove $Size$ from $Sizes$}
 \EndIf

 \State	{$Size = Sizes.Next$}
 \EndIf
  \State
 \EndWhile
\end{algorithmic}
\caption{\label{Alg:FSS} Pseudocode for the fixed set search}
\end{algorithm}

In the pseudocode for the FSS, the first step is initializing the sizes of fixed sets using  \eqref{FixedSizes}.  Next the initial population of solutions is generated performing $N$ iterations of the basic GRASP algorithm. The current size of the fixed set $Size$ is set to the smallest fixed set size. In the main loop, we first randomly generate a set of solutions $\mathcal{S}_{kn}$ by selecting $k$ elements from $\mathcal{P}_n$.  Next, we select a random solution $B$  out of the set $\mathcal{P}_m$. Using $\mathcal{S}_{kn}$, $B$ and $Size$ we generate the fixed set $F$ as described in the previous text. Using $F$ we generate a solution $S= RGF(F)$ using the randomized greedy algorithm with preselected elements. Next, we apply the local search to $S$ and check if we have found a new best solution and add it to the set of generated solutions $\mathcal{P}$. After a new solution is generated we check the two stagnation conditions. The first one checks if the search for the best solution has become stagnant. If so, we set the value of $Size$ to the next value in $Sizes$. Let us note, that the next size is the next larger element of array $Sizes$. In case $Size$ is already the largest size, we select the smallest element in $Sizes$. Before  updating $Size$, we also check if stagnation has occurred in the search of high quality solutions (we have not found a solution which is among the best $n$ or $m$ ones). In case this is true the current $Size$ is removed from $Sizes$. It is important to note, that this is only done if $Size$ is equal to the smallest member of $Sizes$. This is due to the fact that if we have managed to find an improvement for a smaller fixed set than $Size$, it is expected that we have just been ``unlucky" and there is no need to remove this value from the search. This procedure is repeated until the array $Sizes$ is empty or some other termination criterion is satisfied.

\section{Results}

In this section we present the results of our computational experiments used to evaluate the performance of the proposed method. This has been done in a comparison with the GRASP algorithm presented in \cite{Marinakis2005} and the dynamic convexized method (DCTSP) from \cite{7984713}. The focus of the comparison is on the quality of found solutions. 

The FSS and GRASP have been evaluated for both 2-OPT and 3-OPT as local searches. In case of the DCTSP a combination of 2-OPT and 3-OPT has been used as a local search. GRASP has been included to be able to evaluate the effect of the learning mechanism included in the FSS. DCTSP has been used in the comparison since it is a good representative of a metaheuristic whose search is focused on regions near the best solution. In case of the FSS the randomized greedy algorithm used a RCL with 20 elements. In case of both local searches, 2-OPT and 3-OPT, the same size of RCL has been used. To increase the computational efficiency of the local searches we have used the standard approach of ``don't look bits" (DLBs) \cite{RCLDLB}. Let us note, that  when we apply the local search inside the main loop of FSS some of the DLBs could be preset based on the fixed set which significantly decreased the computational cost. The parameters for FSS are the following, the  $k=10$ random  solutions are selected from the best $n=500$ ones for the set of solutions $\mathcal{S}_{kn}$. The base solution is selected from the $m=100$ best solutions. The size of the initial population was $100$. The stagnation criterion was that no new best or high quality solution has been found in the last $Stag=100$ iterations for the current fixed set size. The FSS and GRASP with 2-OPT have been implemented in C\# using Microsoft Visual Studio 2017.  The calculations have been done on a machine with Intel(R) Core(TM) i7-2630 QM CPU \@ 2.00 Ghz, 4GB of DDR3-1333 RAM, running on Microsoft Windows 7 Home Premium 64-bit.

The comparison of the methods has been done on the standard benchmark library TSPLIB \cite{TSPLIB}. The test instances are the same as in \cite{7984713}. A  total of 48 test instances with Euclidean distances are used,  with the number of nodes ranging from 51 to 2392. Note, that in FSS the fact that distances are Euclidean is not exploited. The termination criterion was that a maximal number of solutions has been generated. The limit was the same as in \cite{7984713}, more precisely in case of problem instances having less than a 1000 nodes it was $100|V|$, with $V$ the set of nodes of the considered problem instance,  and in case of larger problem instances it was $10|V|$. For each of the problem instances a single run of each of the methods has been performed, as in \cite{7984713}. The results of the computational experiments can be seen in Table \ref{table:Res}. In it, the results for DCTSP are taken from \cite{7984713} and for GRASP with 3-OPT are taken from \cite{Marinakis2005}. Note, that the results for GRASP-3OPT are very similar to the ones from \cite{Marinakis2005} and to our implementation. 

\begin{table*}[htb]
\footnotesize
\center
\scriptsize
\caption{\label{table:Res}Comparison of the proposed algorithms with GRASP and DCTSP for different TSPLIB instances.}

\begin{tabularx}{420pt}{X*{13}{c}} 

\toprule
 &   \multicolumn{6}{c}{Tour Length}  &&\multicolumn{5}{c}{Relative Error [\%]}\\
  \cmidrule{2-7}   \cmidrule{9-13}
Instance  & \multicolumn{2}{c}{2OPT} &\multicolumn{3}{c}{3OPT} & Known Best &  & 
\multicolumn{2}{c}{2OPT} &\multicolumn{3}{c}{3OPT}\\
  \cmidrule(l){2-3}   \cmidrule(l){4-6}   \cmidrule(l){9-10}   \cmidrule(l){11-13}
   & GRASP & FSS &   GRASP & DCTSP & FSS& &  &  GRASP & FSS &   GRASP & DCTSP & FSS\\

\midrule
eil51 & 426 & 426 & 426 & 426 & 426 & 426 &  & 0.00 & 0.00 & 0.00 & 0.00 & 0.00\\
berlin52 & 7542 & 7542 & 7542 & 7542 & 7542 & 7542 &  & 0.00 & 0.00 & 0.00 & 0.00 & 0.00\\
pr76 & 108351 & 108159 & 108159 & 108159 & 108159 & 108159 &  & 0.18 & 0.00 & 0.00 & 0.00 & 0.00\\
rat99 & 1223 & 1211 & 1211 & 1211 & 1211 & 1211 &  & 0.99 & 0.00 & 0.00 & 0.00 & 0.00\\
kroA100 & 21282 & 21282 & 21282 & 21282 & 21282 & 21282 &  & 0.00 & 0.00 & 0.00 & 0.00 & 0.00\\
kroB100 & 22157 & 22141 & 22141 & 22141 & 22141 & 22141 &  & 0.07 & 0.00 & 0.00 & 0.00 & 0.00\\
kroC100 & 20802 & 20749 & 20749 & 20749 & 20749 & 20749 &  & 0.26 & 0.00 & 0.00 & 0.00 & 0.00\\
kroD100 & 21468 & 21309 & 21294 & 21294 & 21294 & 21294 &  & 0.82 & 0.07 & 0.00 & 0.00 & 0.00\\
kroE100 & 22106 & 22100 & 22068 & 22068 & 22068 & 22068 &  & 0.17 & 0.15 & 0.00 & 0.00 & 0.00\\
rd100 & 7960 & 7910 & 7910 & 7910 & 7910 & 7910 &  & 0.63 & 0.00 & 0.00 & 0.00 & 0.00\\
eil101 & 638 & 629 & 629 & 629 & 629 & 629 &  & 1.43 & 0.00 & 0.00 & 0.00 & 0.00\\
lin105 & 14379 & 14379 & 14379 & 14379 & 14379 & 14379 &  & 0.00 & 0.00 & 0.00 & 0.00 & 0.00\\
pr107 & 44394 & 44303 & 44303 & 44303 & 44303 & 44303 &  & 0.21 & 0.00 & 0.00 & 0.00 & 0.00\\
pr124 & 59159 & 59030 & 59030 & 59030 & 59030 & 59030 &  & 0.22 & 0.00 & 0.00 & 0.00 & 0.00\\
ch130 & 6135 & 6110 & 6110 & 6110 & 6110 & 6110 &  & 0.41 & 0.00 & 0.00 & 0.00 & 0.00\\
pr136 & 98614 & 96920 & 96772 & 96772 & 96772 & 96772 &  & 1.90 & 0.15 & 0.00 & 0.00 & 0.00\\
pr144 & 58554 & 58537 & 58537 & 58537 & 58537 & 58537 &  & 0.03 & 0.00 & 0.00 & 0.00 & 0.00\\
ch150 & 6586 & 6549 & 6528 & 6528 & 6528 & 6528 &  & 0.89 & 0.32 & 0.00 & 0.00 & 0.00\\
kroA150 & 26768 & 26524 & 26524 & 26525 & 26524 & 26524 &  & 0.92 & 0.00 & 0.00 & 0.00 & 0.00\\
pr152 & 74315 & 73682 & 73682 & 73682 & 73682 & 73682 &  & 0.86 & 0.00 & 0.00 & 0.00 & 0.00\\
rat195 & 2391 & 2330 & 2331 & 2323 & 2323 & 2323 &  & 2.93 & 0.30 & 0.34 & 0.00 & 0.00\\
kroA200 & 29803 & 29368 & 29380 & 29382 & 29368 & 29368 &  & 1.48 & 0.00 & 0.04 & 0.00 & 0.00\\
kroB200 & 29909 & 29447 & 29482 & 29437 & 29437 & 29437 &  & 1.60 & 0.03 & 0.15 & 0.00 & 0.00\\
ts225 & 127485 & 127301 & 126643 & 126643 & 126643 & 126643 &  & 0.66 & 0.52 & 0.00 & 0.00 & 0.00\\
pr226 & 80714 & 80369 & 80414 & 80369 & 80369 & 80369 &  & 0.43 & 0.00 & 0.06 & 0.00 & 0.00\\
gil262 & 2456 & 2378 & 2385 & 2379 & 2378 & 2378 &  & 3.28 & 0.00 & 0.29 & 0.04 & 0.00\\
pr264 & 50744 & 49135 & 49135 & 49135 & 49135 & 49135 &  & 3.27 & 0.00 & 0.00 & 0.00 & 0.00\\
a280 & 2658 & 2584 & 2589 & 2579 & 2579 & 2579 &  & 3.06 & 0.19 & 0.39 & 0.00 & 0.00\\
pr299 & 49522 & 48256 & 48235 & 48207 & 48191 & 48191 &  & 2.76 & 0.13 & 0.09 & 0.03 & 0.00\\
rd400 & 15986 & 15322 & 15385 & 15299 & 15284 & 15281 &  & 4.61 & 0.27 & 0.68 & 0.12 & 0.02\\
fl417 & 12066 & 11883 & 11895 & 11883 & 11871 & 11861 &  & 1.73 & 0.19 & 0.29 & 0.19 & 0.08\\
pr439 & 110564 & 107259 & 107401 & 107303 & 107217 & 107217 &  & 3.12 & 0.04 & 0.17 & 0.08 & 0.00\\
pcb442 & 52790 & 50945 & 50946 & 50860 & 50846 & 50778 &  & 3.96 & 0.33 & 0.33 & 0.16 & 0.13\\
d493 & 36192 & 35055 & 35253 & 35136 & 35018 & 35002 &  & 3.40 & 0.15 & 0.72 & 0.38 & 0.05\\
rat575 & 7143 & 6795 & 6863 & 6814 & 6776 & 6773 &  & 5.46 & 0.32 & 1.33 & 0.61 & 0.04\\
p654 & 35113 & 34812 & 34707 & 34658 & 34645 & 34643 &  & 1.36 & 0.49 & 0.18 & 0.04 & 0.01\\
d657 & 51226 & 49258 & 49531 & 49110 & 49014 & 48912 &  & 4.73 & 0.71 & 1.27 & 0.40 & 0.21\\
rat783 & 9352 & 8869 & 8897 & 8848 & 8815 & 8806 &  & 6.20 & 0.72 & 1.03 & 0.48 & 0.10\\
pr1002 & 276251 & 264737 & 262060 & 260218 & 259512 & 259045 &  & 6.64 & 2.20 & 1.16 & 0.45 & 0.18\\
pcb1173 & 61210 & 57788 & 57676 & 57061 & 56965 & 56892 &  & 7.59 & 1.57 & 1.38 & 0.30 & 0.13\\
d1291 & 54537 & 51026 & 51616 & 51099 & 50862 & 50801 &  & 7.35 & 0.44 & 1.60 & 0.59 & 0.12\\
rl1304 & 270441 & 255867 & 255185 & 253842 & 253361 & 252948 &  & 6.92 & 1.15 & 0.88 & 0.35 & 0.16\\
rl1323 & 288538 & 271837 & 273115 & 271914 & 270678 & 270199 &  & 6.79 & 0.61 & 1.08 & 0.63 & 0.18\\
fl1400 & 21044 & 20398 & 20310 & 20167 & 20149 & 20127 &  & 4.56 & 1.35 & 0.91 & 0.20 & 0.11\\
fl1577 & 23274 & 22512 & 22427 & 22352 & 22300 & 22249 &  & 4.61 & 1.18 & 0.80 & 0.46 & 0.23\\
rl1889 & 339151 & 322883 & 319250 & 317825 & 317801 & 316536 &  & 7.14 & 2.01 & 0.86 & 0.41 & 0.40\\
d2103 & 86179 & 81197 & 81312 & 81078 & 80551 & 80450 &  & 7.12 & 0.93 & 1.07 & 0.78 & 0.13\\
pr2392 & 409970 & 387169 & 386017 & 380030 & 379307 & 378032 &  & 8.45 & 2.42 & 2.11 & 0.53 & 0.34\\
\midrule
\multicolumn{7}{l}{Number of found best known solutions}  &  & 2 & 20 & 22 & 27 &30\\
\multicolumn{7}{l}{Average relative error}  &  & 2.73 & 0.40 &0.39  & 0.15 &0.05\\
\bottomrule

\end{tabularx}
\end{table*}

From the results in Table \ref{table:Res}, we can first observe that the GRASP-2OPT has a significantly worse performance than all the other methods finding best known solution for only 4 test instances and having an average relative error of 2.73\%. In case of FSS-2OPT, the improvement is very significant: 20 known best solutions are found and the average relative error was 0.40 \%.  What is very interesting is that FSS-2OPT performs only slightly worse than GRASP-3OPT which finds 22 known best solutions and has an average relative error of 0.39\%. This indicates that the use of the proposed method can be very beneficial in case of less powerful local searches. Although FSS-2OPT had an overall worse performance than DCTSP, it managed to find better solutions for 5 problem instances.  From the results in Table \ref{table:Res}, it is evident that FSS-3OPT has the best performance. It manages to find better solutions than all the other methods for all instances or equal in case methods have found best known solutions. It managed to find 3 more known optimal solutions than DCTSP and has a notable improvement in relative average error. It is important to note that FSS-3OPT never had an error greater than 0.40\%. 

The parameters selected for specifying FSS have been chosen empirically through extensive testing.  Overall the FSS is not highly sensitive to these parameters. The parameter $k$ used to specify the number of solutions selected for  generating the set $S_{kn}$ had the following effect. In case of small values, the selection would result in highly randomized fixed sets. The reason for this is that there are no clear ``good elements", especially in case of larger problem instances where there are not  many common elements in all the solutions. In case of large values of $k$, the method would select very similar fixed sets. The parameter $m$ used to specify the population from which the base solution would be selected has the following effect. In case of small values of $m$ the convergence speed would initially be very fast but would quickly get trapped in locally optimal solutions. This is due to the fact that it becomes very hard to find new high quality solutions. Such values of $m$ are useful in case we can only generate a small number of solutions. In case of high values of $m$, the convergence speed is much slower. The problem  is that when the fixed set is generated for a lower quality solution $B$, although the method manages to find solutions of higher quality than $B$, it is unlikely that very high quality ones will be found. The effect of parameter $n$ for specifying the size of the population used for generating $S_{kn}$ had a similar effect but to a much lower extent. We found that the best choice of parameters for stagnation, both for finding best and high quality solutions, was the same as the number of iterations from which GRASP would rarely find new best solutions. In general, it is important to avoid very small values for this parameter because it results in prematurely stopping the evaluation of small fixed sets. 

\begin{figure}[ht]
\centering
%\subfigure[pr264]{%
\includegraphics[width=0.75\textwidth]{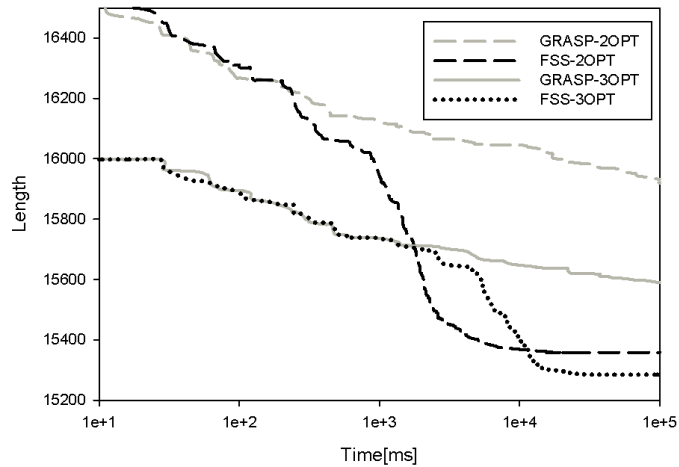}
%\label{fig:subfigure1}}
%\quad
%\subfigure[pr1002]{%
%\includegraphics[width=0.45\textwidth]{pr1002}
%\label{fig:subfigure4}}
%
\caption{Average solution quality of 10 independent  runs of FSS and GRASP for the TSPLIB problem instance {\it rd400}.}
\label{fig:pr264}
\end{figure}
\begin{figure}[ht]
\centering
%\subfigure[pr264]{%
%\includegraphics[width=0.75\textwidth]{pr264}
%\label{fig:subfigure1}}
%\quad
%\subfigure[pr1002]{%
\includegraphics[width=0.75\textwidth]{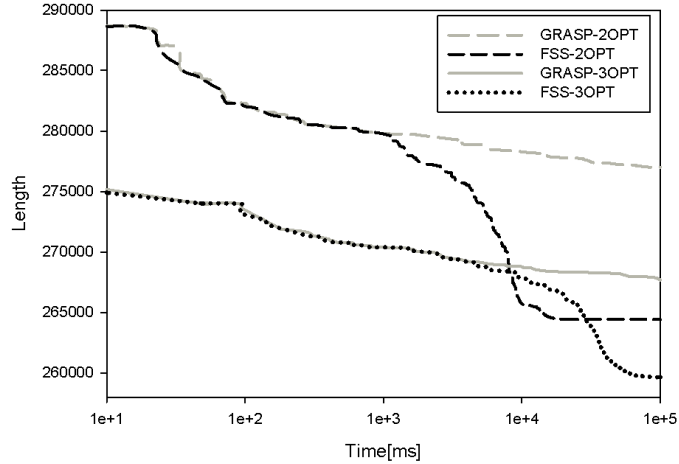}
%\label{fig:subfigure4}}
%
\caption{Average solution quality of 10 independent  runs of FSS and GRASP for the TSPLIB problem instance {\it pr1002}.}
\label{fig:pr1002}
\end{figure}

In Table \ref{table:Res} we did not include computational times, since they are highly dependent on structures used for implementing the local searches. We would like to note that in case of our implementation of FSS and GRASP there was a significant decrease in computational time. This is due to the fact of a smaller candidate set for the greedy algorithm. The second reason is that  the number of iterations needed to generate the solution was significantly lower. As it is well known, the computational cost of 2-OPT and 3-OPT local searches is significantly higher than for generating the initial solution. An extensive analysis of the computational cost of 2-OPT can be found in \cite{Englert2014}. In case of FSS, we exploit the fact that the fixed set is a subset of a locally optimal base solution $B$ through ``don't look bits". More precisely, the DLBs of all the inner points of paths in the fixed set can be preset. Note that, the number of preset DLBs is close to the size of the fixed set. In practice this means instead of having the computational cost of the first iteration of 2-OPT being proportional to $|V|C$, were $C$ is the size of  the RCL, it is close to  $(|V| -|F|) C$. Similar analysis can be done for the 3-OPT local search. In case of very large fixed sets, the time FSS generated a new solution and applied the local search was a fraction of the time need to accomplishing the same task in GRASP.  It is expected that a similar behavior would be present in applying FSS to other combinatorial problems. 

This decrease in computational cost of the FSS compared to GRASP directly effects the convergence speed.  This is illustrated in Fig. \ref{fig:pr264}, \ref{fig:pr1002} for representative problem instances. In each of the figures the convergence speed of the average solution length for ten independent runs of GRASP and FSS, with 2-OPT and 3-OPT used as local searches, are shown. It can be observed that there is a drastic increase in the convergence speed after the initial population is generated for the FSS.

\section{Conclusion}

In this paper we have presented a new metaheuristic called fixed set search that exploits the common elements of high quality solutions. The proposed metaheuristic represents a method of adding a learning mechanism to the GRASP metaheuristic. It is expected that FSS can be applied to a wide range of problems since the only requirement is that the solution can be represented in a set form. A very important aspect of FSS is the simplicity in which a GRASP algorithm can be adapted to it. This is done with two basic steps. Firstly, the  randomized greedy algorithm, used in the GRASP, is adapted to a setting were some elements are preselected. We have shown that this can be trivially achieved in case of the TSP, and it is expected that this is the case for many other combinatorial problems. Secondly, the method for generating a fixed set needs to be implemented which consists in selecting several solutions and tracking the number of times their elements occur in a selected base solution. 
 
  We have illustrated the effectiveness of the proposed approach on the TSP. Our computational experiments have shown that the proposed method has a significantly better performance than the basic GRASP approach when both solution quality and convergence speed are considered. Further, we have shown that the approach has a considerably better performance than the  dynamic convexized method applied to the TSP in case 3-OPT is used as a local search. The proposed method has proven very efficient in improving the performance of GRASP in case of a less powerful local search. 
  
  It is important to note that there is a wide range of potential improvements to the proposed method. Some examples are having a more intelligent method of selecting the solutions used in generating the fixed set; or adapting the computational effort used to solve the subproblem related to a specific fixed set. Our objective was to show that even in the most basic form the proposed method can produce a significant improvement. We would like to note that the concept of using a fixed set can potentially be used to hybridize other metaheuristics like ACO,  genetic algorithms and similar, for solving large scale problems through focusing the search in some promising areas of the solution space. On the other hand the idea of fixing elements of a solution can easily be included in mixed integer programs so there is a potential of adapting FSS to a matheuristic setting. 

%
% ---- Bibliography ----
%
% BibTeX users should specify bibliography style 'splncs04'.
% References will then be sorted and formatted in the correct style.
%

\end{document}